\documentclass[preprint,12pt]{elsarticle}

\usepackage{amssymb}
\usepackage{amsmath}
\usepackage{amsthm}

\usepackage{multirow}
\usepackage{adjustbox}
\usepackage{booktabs}
\usepackage{xurl}

\usepackage[utf8]{inputenc}

\usepackage{hyperref}
\hypersetup{
	unicode,
	pdfauthor={Yanchen Wang, Adam Turnbull, Tiange Xiang, Yunlong Xu, Sa Zhou, Adnan Masoud, Shekoofeh Azizi, Feng Vankee Lin, Ehsan Adeli},
	pdftitle={Decoding Visual Experience and Mapping Semantics through Whole-Brain Analysis Using fMRI Foundation Models},
	pdfsubject={fMRI Foundation Models for Vision Decoding},
	pdfkeywords={fMRI, vision decoding, foundation models, whole-brain analysis},
	pdfproducer={LaTeX},
	pdfcreator={pdflatex}
}

\biboptions{sort&compress}

\def\citeauthor#1{\NAT@cite{#1}{}{}{}{}}

\theoremstyle{plain}

\theoremstyle{definition}

\newcommand{\R}{\mathbb{R}}

\usepackage{color}
\graphicspath{{fig/}}

\usepackage{algorithm, algpseudocode}

\usepackage{mathrsfs}

\usepackage{lipsum}

\journal{Medical Image Analysis}

\usepackage{color}
\definecolor{MyDarkBlue}{rgb}{0,0.08,0.8} 
\definecolor{MyBlack}{rgb}{0,0,0}
\newcommand{\blue}[1]{\textcolor{MyBlack}{#1}}

\begin{document}

\begin{frontmatter}



\title{Decoding Visual Experience and Mapping Semantics through Whole-Brain Analysis Using fMRI Foundation Models}


\author[aff1]{Yanchen Wang}
\ead{ppwang@cs.stanford.edu}

\author[aff1]{Adam Turnbull}
\ead{aturnbu2@stanford.edu}

\author[aff2]{Tiange Xiang}
\ead{xtiange@stanford.edu}

\author[aff3]{Yunlong Xu}
\ead{dracoxu@uchicago.edu}

\author[aff1]{Sa Zhou}
\ead{sazhou@stanford.edu}

\author[aff4]{Adnan Masoud}
\ead{Adnan.Masood@ust.com}

\author[aff5]{Shekoofeh Azizi}
\ead{shekazizi@google.com}

\author[aff1]{Feng Vankee Lin\fnref{fn2}}
\ead{fvlin@stanford.edu}

\author[aff1,aff2]{Ehsan Adeli\fnref{fn2}\corref{cor1}}
\ead{eadeli@stanford.edu}

\affiliation[aff1]{organization={Department of Psychiatry and Behavioral Sciences, Stanford University},
            city={Stanford},
            state={CA},
            country={USA}}

\affiliation[aff2]{organization={Department of Computer Science, Stanford University},
            city={Stanford},
            state={CA},
            country={USA}}

\affiliation[aff3]{organization={Department of Neurobiology, University of Chicago},
            city={Chicago},
            state={IL},
            country={USA}}

\affiliation[aff4]{organization={UST},
            city={Tampa},
            state={FL},
            country={USA}}

\affiliation[aff5]{organization={Google DeepMind},
            city={Toronto},
            state={Ontario},
            country={Canada}}

\fntext[fn2]{Senior authors}
\cortext[cor1]{Corresponding authors}

\begin{abstract}

Neural decoding, the process of understanding how brain activity corresponds to different stimuli, has been a primary objective in cognitive sciences. Over the past three decades, advances in functional Magnetic Resonance Imaging (fMRI) and machine learning have greatly improved our ability to map visual stimuli to brain activity, especially in the visual cortex. Concurrently, research has expanded to decode more complex processes, such as language and memory across the whole brain, using techniques to handle greater variability and improve signal accuracy. We argue that ``seeing" involves more than just mapping visual stimuli onto the visual cortex; it engages the entire brain, as various emotions and cognitive states can emerge from observing different scenes. In this paper, we develop algorithms to enhance our understanding of visual processes by incorporating whole-brain activation maps while individuals are exposed to visual stimuli. We utilize transformer-based large-scale fMRI encoders and Image generative models (encoders \& decoders) pre-trained on large public datasets, which are then fine-tuned through Image-fMRI contrastive learning. Our models can decode visual experience across the entire cerebral cortex, surpassing the traditional confines of the visual cortex. Using a public dataset (BOLD5000), we first compare our method with state-of-the-art approaches for decoding visual processing and show improved predictive semantic accuracy by \textbf{43\%}. A network ablation analysis suggests that, beyond the visual cortex, the default mode network contributes significantly to stimulus decoding, in line with the proposed role of this network in sense-making and semantic processing. Additionally, we implemented zero-shot (i.e., no additional training) imagination decoding on an independent validation dataset, achieving a statistically significant ($p$-value $< 0.05$) mapping between the reconstructed images and ground-truth text stimuli. This demonstrates the model's generalization capability to capture semantic meanings across various scenarios. These findings underscore the potential of employing comprehensive models to enhance the nuanced interpretation of semantic processes.
\end{abstract}


\end{frontmatter}


\section{Introduction}
\label{sec:intro}

Understanding the relationship between brain activity and the characteristics of external stimuli, commonly known as ``neural decoding" \citep{kamitani2005decoding} or identifying ``neural representations," \citep{kriegeskorte2019peeling} is fundamental to advancing neuroscience. 
In the last three decades, the development and continuous improvement of functional magnetic resonance imaging (fMRI) has allowed researchers to map spatial patterns of activity at the level of millimeter-cubed ``voxels" onto patterns of stimuli \blue{to discern how specific information is encoded}. \blue{The majority of research has focused on mapping representations of visual stimuli onto the visual cortex with increasing success~\cite{kamitani2005decoding,takagi2023high,chen2023seeing,scotti2024reconstructing,xia2024umbrae}}.

The field of visual decoding has advanced steadily in terms of the accuracy with which stimuli can be predicted from brain activity. \blue{This progress has accelerated} with the recent improvements in machine learning and deep learning methods that can identify more complex \blue{non-linear} relationships between visual activity and visual stimuli~\citep{chen2023seeing,scotti2024reconstructing,takagi2023high,ozcelik2023natural}.

While recent advances in neural decoding have significantly enhanced the ability \blue{to reconstruct visual stimuli from the visual cortex}, these models predominantly excel at capturing basic, low-level visual features such as edges, textures, and other pixel-wise attributes~\citep{kamitani2005decoding,beliy2019voxels}. \blue{However, human vision is a holistic process where sensory input is integrated with memory and emotion~\cite{pessoa2010emotion,khan2011visual}.} 

Running parallel to this work have been attempts to map the representations of more complex higher-order processes such as language~\citep{anderson2019pattern}, semantics~\citep{huth2016natural}, movie-watching~\citep{guntupalli2016model}, or autobiographical memory~\citep{anderson2020decoding}. \blue{Unlike localized visual decoding, these studies demonstrate that information is often encoded across broad cortical networks~\cite{power2017sources}.} \blue{For instance, seminal work by Huth et al.~\citep{huth2016natural} revealed that semantic concepts are mapped systematically across the cerebral cortex, suggesting that ``meaning" is not confined to a single region. However, because these distributed patterns exhibit high inter-subject variability, traditional statistical approaches, such as multi-voxel pattern analysis (MVPA)~\citep{norman2006beyond} or representational similarity analysis (RSA)~\citep{kriegeskorte2008representational}, often require datasets with a large amount of participants, taking advantage of individual differences to understand shared variance.} This can be contrasted with visual decoding that typically uses longer scans in fewer individuals to capture basic visual processes that vary less across people. 

Recent evidence increasingly \blue{implicates the whole-brain in fundamental visual tasks.} \blue{Notably, studies suggest that the default mode network~(DMN) may maintain a reinotopic code to bridge past experiences with incoming visual} information, \blue{while affordance-based representations of objects appear distributed globally ~\citep{steel2024retinotopic,thornton2024neural}}. \blue{Theories of predictive coding~\citep{aitchison2017or} and active inference ~\citep{friston2009reinforcement} further argue that visual perception relies on top-down comparisons with internal world models housed in higher-order regions}. 
\blue{Despite this theoretical motivation,} expanding \blue{visual decoding} to the whole-brain \blue{presents a severe computational bottleneck due to the type of data typically used for these tasks." Whole-brain data introduces a feature space that vastly outscales the number of available training samples and the lack of between-subject variance in less basic visual processes (e.g., those involving subjective experience). Previous methods circumvented this by restricting analyses to the visual cortex, allowing simpler linear models to function effectively within a constrained search space~\cite{takagi2023high,kamitani2005decoding} but only allowing the modeling of basic processes that are similar across participants. To decode more complex, variable visual processes from the whole brain without overfitting, we therefore require models with strong inductive priors and the capacity to transfer robust between-subject representations to high-dimensional, noisy inputs.}

\blue{To address this challenge, we introduce WAVE (Whole-brain Analysis of Visual Experience), a framework that }leverages \blue{the capacity of foundation models~\citep{bommasani2021opportunities} to extract universal neural dynamics from large-scale generic data. By transferring these learned priors to the visual decoding domain, WAVE can effectively model high-dimensional whole-brain representations even in datasets with a limited number of participants~}(Fig.~\ref{fig:figure_1}). \blue{Our contribution is threefold.} First, we compare our method with state-of-the-art approaches to decoding visual processing and show improved predictive accuracy. We highlight that even after removing the visual cortex we can show high predictive accuracy on a traditional visual dataset, emphasizing the importance of whole-brain processing for even simple visual tasks. Subsequently, we enhance our model's interpretability through post-hoc analyses. Our findings reveal that the higher-order cortical hierarchy provides more predictive power for visual scenes with greater complexity. Furthermore, we demonstrate the model's ability to capture semantic information across different scenarios. We used a separate fMRI imagination dataset where participants gave verbal descriptions followed by fMRI during scenario visualization. Our model excelled at decoding scenarios semantically similar to images from the original dataset. This highlights \blue{WAVE's} capacity to generalize semantic understanding beyond the original training data. This research represents both methodological and conceptual advances: we believe that \blue{our method} can improve the accuracy of brain-stimuli mapping in a way that can be applied across a greater range of datasets. Ultimately, this could lead to more precise modeling of \blue{distributed} neural processes, gaining novel insights into human \blue{cognition.}
\section{Results}
\begin{figure}[t!]
    \centering
    \includegraphics[width=\linewidth]{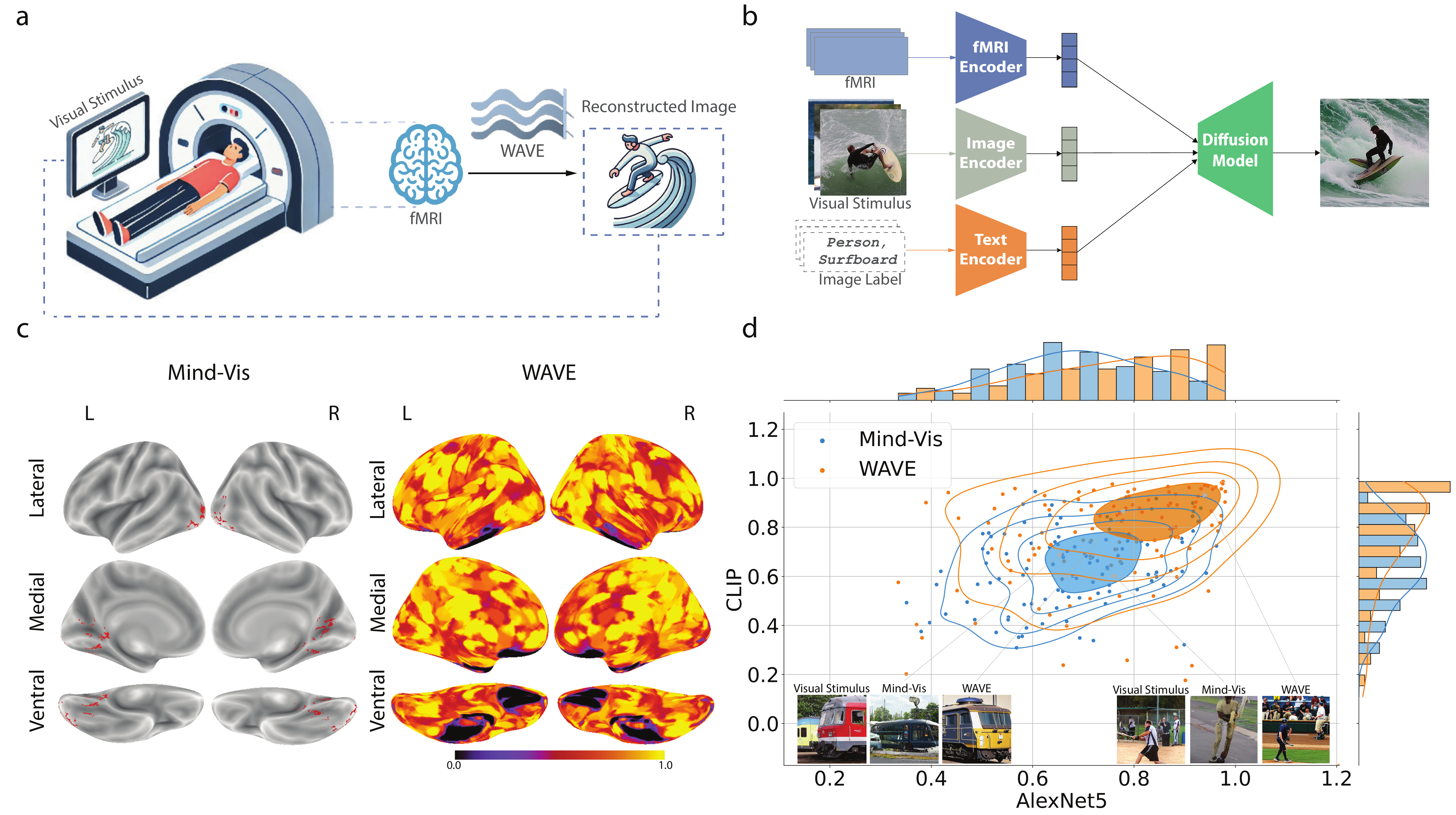}
    \caption{\textbf{Model Pipeline and performance comparison with previous work.} \textbf{a,} Study Formation: The diagram illustrates our model, WAVE, which reconstructs visual stimuli from fMRI data. \textbf{b,} Model Training Framework: After preprocessing the raw fMRI data, WAVE integrates three modalities: fMRI, image, and text to perform contrastive learning. The features are then passed to a diffusion model for final image reconstruction. \textbf{c,} Brain Saliency Maps: The saliency maps compare the input fMRI data between Mind-Vis and WAVE. Mind-Vis utilizes only visual cortex data (highlighted \blue{voxels} in red), whereas WAVE employs whole-brain fMRI data. The saliency map for WAVE is based on the model's attention, demonstrating a broader and more comprehensive engagement of brain regions. \textbf{d,} Quantitative Analysis: This graph compares the performance of the universal model in reconstructing images between Mind-Vis and WAVE. The y-axis represents CLIP accuracy~\citep{radford2021learning} from high-level metrics, and the x-axis represents AlexNet(5)~\citep{krizhevsky2012imagenet} accuracy from low-level metrics~\citep{ozcelik2023natural}. The distributions are displayed on the sides, along with sample images, \blue{showing the superior performance} of WAVE in capturing both high-level and low-level features.}
    \label{fig:figure_1}
\end{figure}
In this study, we explored human visual experiences through fMRI, leveraging an \blue{architecture that integrates a transformer-based} fMRI foundation model \blue{with a latent diffusion} generative model. \blue{While prior studies~\cite{scotti2024reconstructing,chen2023seeing,sun2024contrast,xia2024umbrae}} predominantly focused on the visual cortex, which is already known for its role in visual processing~\cite{huff2018neuroanatomy,hubel1964effects}, \blue{this whole-brain approach allowed us to generate attention-based saliency maps that reveal how semantic information is encoded across diverse brain networks beyond early visual areas~(Fig.~\ref{fig:figure_1}c).} We employed a range of metrics to evaluate the reconstructed images produced by our model, emphasizing semantic accuracy~\cite{chen2023seeing}, high-level and low-level correlations~\cite{ozcelik2023natural}. 

Going beyond the individual model reconstruction typical of previous work~\cite{chen2023seeing,scotti2024reconstructing,sun2024contrast}, we extended our experiment to include a universal setting, wherein a single model decodes neural activities across different subjects. We conducted a qualitative comparison of our approach with another advanced method Mind-Vis in Fig.~\ref{fig:figure_1}d. We also conducted extensive ablation studies, demonstrating that whole-brain data offers greater semantic advantages over data solely from the visual cortex. 

Finally, we analyzed various clusters of images and presented \blue{brain }saliency maps from our model, \blue{improving }the explainability of our \blue{method} and providing insights through cognitive neuroscience statistical analysis.

\subsection{Decoding images from whole-brain fMRIs}
We successfully decoded images from whole-brain fMRI data using two distinct methods: building individual models for each subject, as done in visual decoding studies such as~\cite{chen2023seeing}, and constructing a universal model applicable across four subjects (details in Section~\ref{method:dataset}). We evaluated our model's performance by comparing the reconstructed images against the original visual stimuli, the ground-truth, using a range of metrics that examined the semantic meaning, high-level relationships, and low-level pixel-wise distances (details in Section~\ref{method:metrics}). 

\begin{figure}[!ht]
    \centering
    \includegraphics[width=\linewidth]{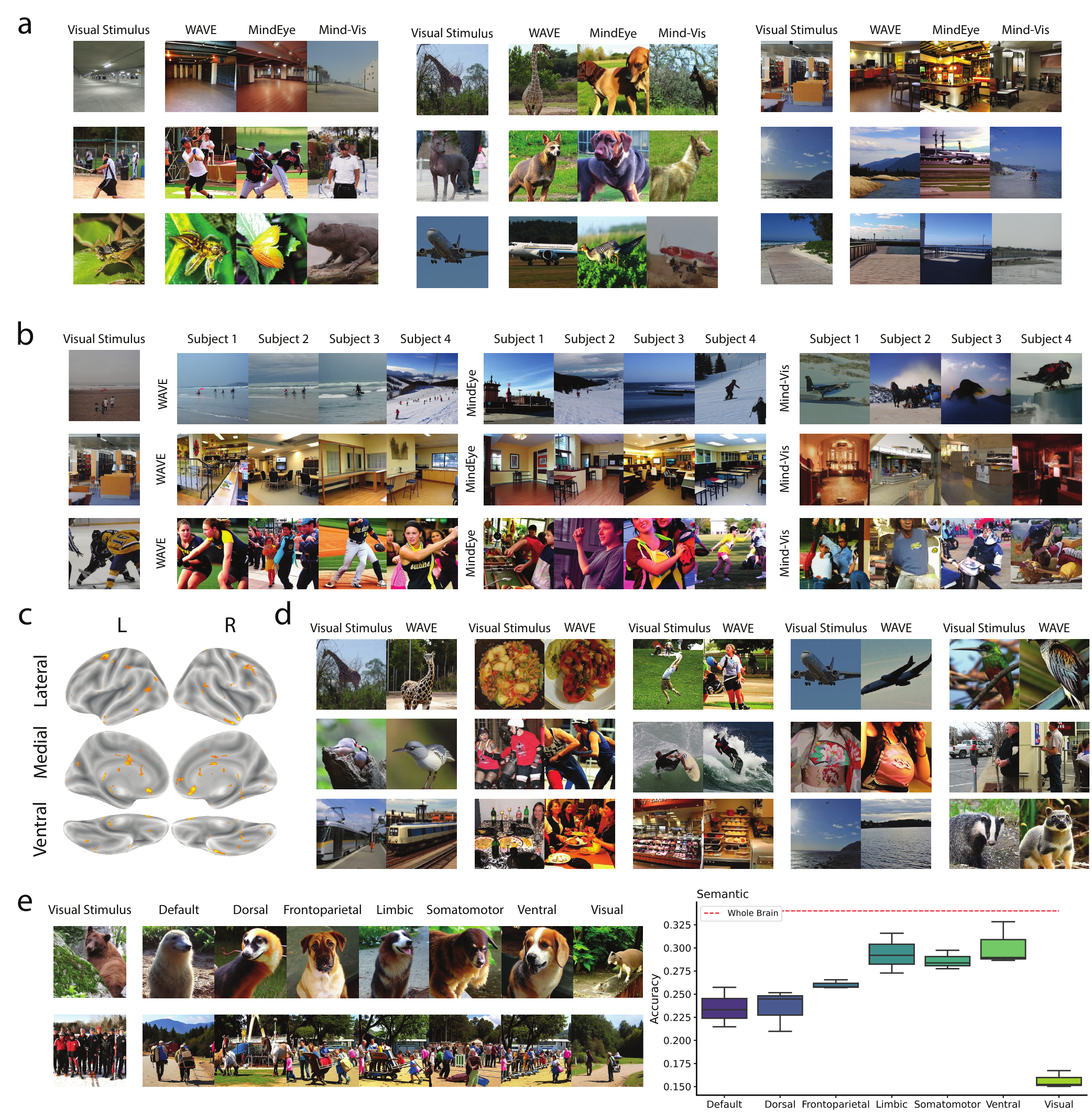}
    \vspace{-20pt}
    \caption{\scriptsize\blue{\textbf{Comparative Evaluation and Network Ablation of the WAVE Framework.}} a, \blue{Qualitative comparison of reconstructed images. Columns display the original visual stimulus alongside reconstructions from WAVE and baselines}. b, Universal settings across all four subjects in the BOLD5000 dataset. This panel demonstrates the generalization capability of WAVE, MindEye, and Mind-Vis among different subjects. c, The saliency map of the WAVE model using data where the visual cortex has been masked, showing the top 20 regions of interests. d, Decoding from the visual cortex-masked model: examples of reconstructed images using non-visual fMRI regions. In each pair, the left image shows the original visual stimulus, and the right image shows the reconstructed image generated by our visual cortex-masked method (WAVE). e, Impact of network ablation on Decoding Accuracy (subject CSI-1). The left side displays generated images resulting from the masking of each of the seven networks. The right side features a box plot illustrating the decoding accuracy for each network ablation, highlighting the accuracy reduction when specific networks are removed. The red dashed line above represents the whole-brain decoding accuracy for comparison.}
    \label{fig:vis_mask_examples}
\end{figure}

As shown in Table~\ref{tab:table_1}, we benchmarked our model, WAVE, with best in class models MindEye~\cite{scotti2024reconstructing} and MindVis~\cite{chen2023seeing}. The results of individual subject-based models are detailed in \blue{Table~\ref{tab:table_4}}. In both individual and universal settings, WAVE achieved the best performance in Semantic Top-1 Accuracy and high-level metrics. In low-level metrics, WAVE also outperformed AlexNet(2) and AlexNet(5). Notably, while SSIM and PixCorr metrics measure pixel-wise correlations, all other metrics from both low-level and high-level categories measure encoded image features from models~(e.g. AlexNet, CLIP). (See metrics details in Section~\ref{method:metrics}) This suggests that our model excels in semantic and high-level image reconstruction. We also provide rich reconstructed image examples comparing WAVE with MindEye and Mind-Vis in Fig.~\ref{fig:vis_mask_examples}a. 

Furthermore, the results from the universal decoding setting, as shown in Table~\ref{tab:table_2} underscore WAVE's superior performance and the advantage of utilizing whole-brain data compared to Mind-Vis~\cite{chen2023seeing}. Due to individual variability, brain fMRI data differ significantly between voxels. The same registered voxel in standard space usually does not have the same function across subjects. However, brain regions tend to maintain similar functions across individuals. Our brain-wide network preprocessed fMRI data showed more advantages in universal settings. Additionally, by leveraging the pre-trained foundation fMRI model, WAVE demonstrated better generalization power, highlighting the robustness and adaptability of our approach. The reconstructed examples in Fig.~\ref{fig:vis_mask_examples}b clearly show our model's consistency of the generated image across four subjects. WAVE can capture the semantic meaning from brain-wide network activity accurately compared with other methods.

\begin{table}[t]
    \caption{\textbf{An evaluation of our model's capability to reconstruct images from fMRI data.} Employing Semantic metrics as~~\cite{chen2023seeing}, we adopt a 50-way top-1 accuracy approach~(chance=2\%) over 1000 trials to gauge the fidelity of reconstructed images against the original stimuli. We used a similar setting to previous study, using Low-Level and High-Level metrics. We used pixel-wise correlation~(PixCorr) and structural similarity index metrics~(SSIM). Additionally, we assess correlation distance using the EfficientNet-B1~(Eff)~\citep{tan2019efficientnet} and SwAV-ResNet50~(SwAV)~\citep{caron2020unsupervised} models. For other metrics, two-way identification~(chance=50\%) is used for evaluation, consistent with methodologies described by~\cite{ozcelik2023natural,scotti2024reconstructing}. Two experimental setups were investigated: one involving individualized models for each subject, with an average taken across four subjects' metric results (detailed results in the appendix), and another employing a universal model for all subjects.}
    \begin{adjustbox}{width=\columnwidth}
    \begin{tabular}{l|c|cccc|cccc|c}
    \toprule
    \multirow{2}{*}{\textbf{Setting}} &
      \multirow{2}{*}{\textbf{Model}} &
      \multicolumn{4}{c|}{\textbf{Low-Level}} &
      \multicolumn{4}{c|}{\textbf{High-Level}} &
      \textbf{Semantic} \\ 
     &          & PixCorr↑ & SSIM↑          & AlexNet(2)↑ & AlexNet(5)↑ & Incep↑  & CLIP↑   & Eff↓  & SwAV↓ & Top-1 Acc↑ \\ \hline
    \multirow{3}{*}{Individual} &
      WAVE~(Ours) &
      0.062 &
      0.199 &
      \textbf{71.61\%} &
      \textbf{79.70\%} &
      \textbf{68.76\%} &
      \textbf{81.54\%} &
      \textbf{0.892} &
      \textbf{0.552} &
      \textbf{25.24\%} \\
     & MindEye~\citep{scotti2024reconstructing}  & 0.056    & 0.217          & 71.11\%     & 77.31\%     & 66.32   & 69.98\% & 0.916 & 0.592 & 16.85\%   \\
     & Mind-Vis~\citep{chen2023seeing} & \textbf{0.074}    & \textbf{0.308} & 67.42\%     & 72.74\%     & 66.97\% & 70.07\% & 0.914 & 0.556 & 17.66\%   \\ 
     \hline
    \multirow{2}{*}{Universal} &
      WAVE~(Ours) &
      \textbf{0.050} &
      0.194 &
      \textbf{69.47\%} &
      \textbf{78.31\%} &
      \textbf{68.41\%} &
      \textbf{78.41\%} &
      \textbf{0.902} &
      \textbf{0.591} &
      \textbf{20.75\%} \\
     & Mind-Vis~\citep{chen2023seeing} & 0.036    & \textbf{0.272} & 60.56\%     & 68.92\%     & 64.25\% & 64.47\% & 0.938 & 0.593 & 9.19\%    \\ 
     \bottomrule
    \end{tabular}
    \label{tab:table_1}
    \end{adjustbox}

\end{table}

\paragraph{Ablation Study}
We conducted an ablation study to evaluate the performance of our model using different types of fMRI data. The base model used in this study was MindEye, which differs from Mind-Vis in that it does not utilize a pre-trained voxel encoder. Additionally, MindEye employs a Residual MLP Backbone, making it easier to adjust input dimensionality.

As shown in Table~\ref{tab:table_2}, whole-brain network data demonstrated significantly better semantic characteristics compared to data derived from visual cortex voxels. In both individual and universal settings, whole-brain fMRI data achieved state-of-the-art results across all high-level metrics. While PixCorr and SSIM metrics relate more to visual cortex area features, the whole-brain data incorporates complementary information from high-hierarchy functional brain regions involved in visual processing. This comprehensive approach highlights the semantic advantages of using whole-brain data for image reconstruction. \blue{Furthermore, additional analyses on the encoder's representational capacity (Table~\ref{tab:encoder_retrieval} and Table~\ref{tab:subject_classification}) suggest that the pre-trained fMRI model plays a pivotal role in effectively extracting and integrating these complex whole-brain signals.}

\begin{table}[ht!]
\caption{\textbf{Ablation study results comparing whole-brain and visual cortex data using the MindEye model.} The table illustrates that whole-brain data significantly outperforms visual cortex data in semantic metrics across both individual and universal settings. High-level metrics also show superior performance for whole-brain data, indicating that it captures more comprehensive and high-hierarchy functional information for visual processing. While PixCorr and SSIM metrics, related to visual cortex features, are slightly lower for whole-brain data, the overall advantages in semantic and high-level metrics highlight the robustness and effectiveness of using whole-brain fMRI data for model decoding.}
\begin{adjustbox}{width=\columnwidth}
\begin{tabular}{c|c|cccc|cccc|c}
\toprule
\multirow{2}{*}{\textbf{Setting}} &
  \multirow{2}{*}{\textbf{Data}} &
  \multicolumn{4}{c|}{\textbf{Low-Level}} &
  \multicolumn{4}{c|}{\textbf{High-Level}} &
  \textbf{Semantic} \\ 
 &
   &
  PixCorr$\uparrow$ &
  SSIM$\uparrow$ &
  AlexNet(2)$\uparrow$ &
  AlexNet(5)$\uparrow$ &
  Incep$\uparrow$ &
  CLIP$\uparrow$ &
  Eff$\downarrow$ &
  SwAV$\downarrow$ &
  Accuracy$\uparrow$ \\ \hline
\multirow{2}{*}{Indvidual} &
  Visual Cortex &
  0.056 &
  \textbf{0.217} &
  71.11\% &
  77.31\% &
  66.32\% &
  68.11\% &
  \textbf{0.916} &
  0.592 &
  16.85\% \\
 &
  Whole Brain &
  \textbf{0.084} &
  0.215 &
  \textbf{71.48\%} &
  \textbf{79.49\%} &
  \textbf{68.64\%} &
  \textbf{76.29\%} &
  0.908 &
  \textbf{0.579} &
  \textbf{19.60\%} \\ \hline
\multirow{2}{*}{Universal} &
  Visual Cortex &
  0.081 &
  \textbf{0.214} &
  68.32\% &
  73.82\% &
  63.31\% &
  69.75\% &
  0.929 &
  0.609 &
  14.43\% \\
 &
  \multicolumn{1}{l|}{Whole Brain} &
  \textbf{0.091} &
  0.213 &
  \textbf{68.96\%} &
  \textbf{76.67\%} &
  \textbf{66.60\%} &
  \textbf{76.18\%} &
  \textbf{0.903} &
  \textbf{0.596} &
  \textbf{18.72\%} \\ 
\bottomrule
\end{tabular}
\end{adjustbox}
\label{tab:table_2}
\end{table}


\paragraph{Decoding Visual Stimuli Beyond the Visual Cortex}
\blue{Our research is guided by the hypothesis that neural representations of visual experience are distributed well beyond the primary visual cortex.} We posit that \blue{non-visual regions encode essential high-level} semantic aspects of visual stimuli, \blue{which are crutial for accurate semantic-correct image reconstruction}. \blue{To validate this, we performed a series of \blue{computational lesioning} experiments to isolate the contributions of specific brain networks.}

\blue{First, we assessed the model's capacity to \textbf{decode visual stimuli solely from non-visual regions} by completely masking the visual network. As illustrated in Figure~\ref{fig:vis_mask_examples}d, the resulting reconstructions retained remarkable semantic fidelity, achieving an average semantic accuracy of \textbf{18.98\%} across four subjects. While this is lower than the \textbf{25.24\%} accuracy achieved by the whole-brain model (Table~\ref{tab:table_1}), the relatively high performance confirms that non-visual networks contain sufficient information to drive semantic decoding. The saliency map highlighting the top 20 non-visual regions contributing to this process is displayed in Figure~\ref{fig:vis_mask_examples}c.}

\blue{To further dissect these contributions, we conducted a \textbf{systematic network ablation study} using the Yeo 7-network parcellation~\citep{yeo2011organization}. We iteratively masked one network at a time while using the remaining six for decoding. Results for subject CSI-1 (Figure~\ref{fig:vis_mask_examples}e) indicate that while removing the visual network caused the most significant drop in accuracy (marked by the red dashed line), the exclusion of the \blue{Default Mode Network (DMN) and Dorsal Attention Network} also resulted in substantial performance decrements. This suggests these networks are not merely auxiliary but \blue{integral} to the decoding process. This finding aligns with established neuroscientific literature: the dorsal attention network supports higher-order visual attention, while the DMN is implicated in integrating semantic context and ``making sense" of sensory inputs~\cite{yeshurun2021default}.}

\subsection{Post-hoc analysis of Whole Brain Visual Experience }
\begin{figure}
    \centering
    \includegraphics[width=\linewidth]{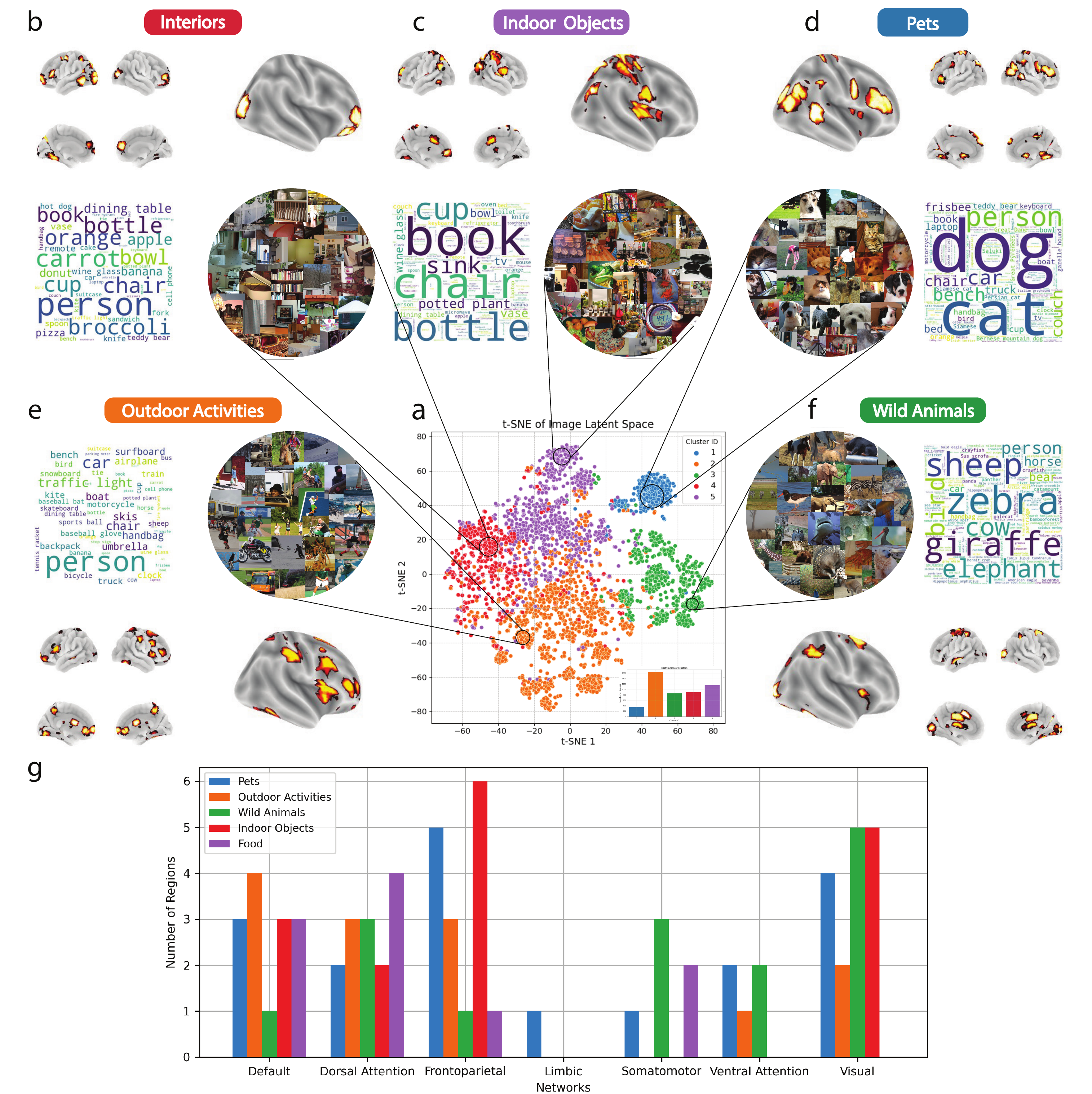}
    \vspace{10pt}
    \caption{\blue{\textbf{Semantic Profiling of Whole-Brain Visual Representations.} }
    a, t-SNE \blue{projection} of image embeddings, \blue{colored by five distinct semantic clusters identified via K-Means.} b-f, \blue{Detailed profiles for each cluster.} Word clouds \blue{illustrate the most frequent object labels within each group} The subfigure titles, generated by entering the words into ChatGPT-4, summarize the thematic essence of each cluster. Accompanying each word cloud are selected image samples and a whole-brain saliency map highlighting the top 20 regions of interest relevant to the cluster. g, \blue{Network-level decomposition of the saliency maps. The bar chart quantifies quantifies the distribution of top regions across Yeo-7}, providing insights into the network-based localization of visual processing associated with different categories.}
    \label{fig:k-cluster}
\end{figure}

To \blue{investigate how distinct visual concepts recruit specific whole-brain networks}, we performed an extensive post-hoc \blue{semantic profiling }analysis of whole-brain visual experiences using our WAVE model. \blue{We applied unsupervised clustering to the stimulus space, partitioning} original images into five distinct \blue{semantic} groups and analyzed the corresponding patterns of brain activity. As illustrated in Fig.~\ref{fig:k-cluster}, the brain saliency maps highlight the regions of interest (ROIs) activated by each image cluster. These findings reveal that different clusters are associated with distinct neural activation patterns, but also demonstrate that similar brain networks were important for decoding across different clusters. 

\subsection{Zero-Shot Human Imagination Decoding}
\begin{figure}[t!]
    \centering
    \includegraphics[width=0.9\linewidth]{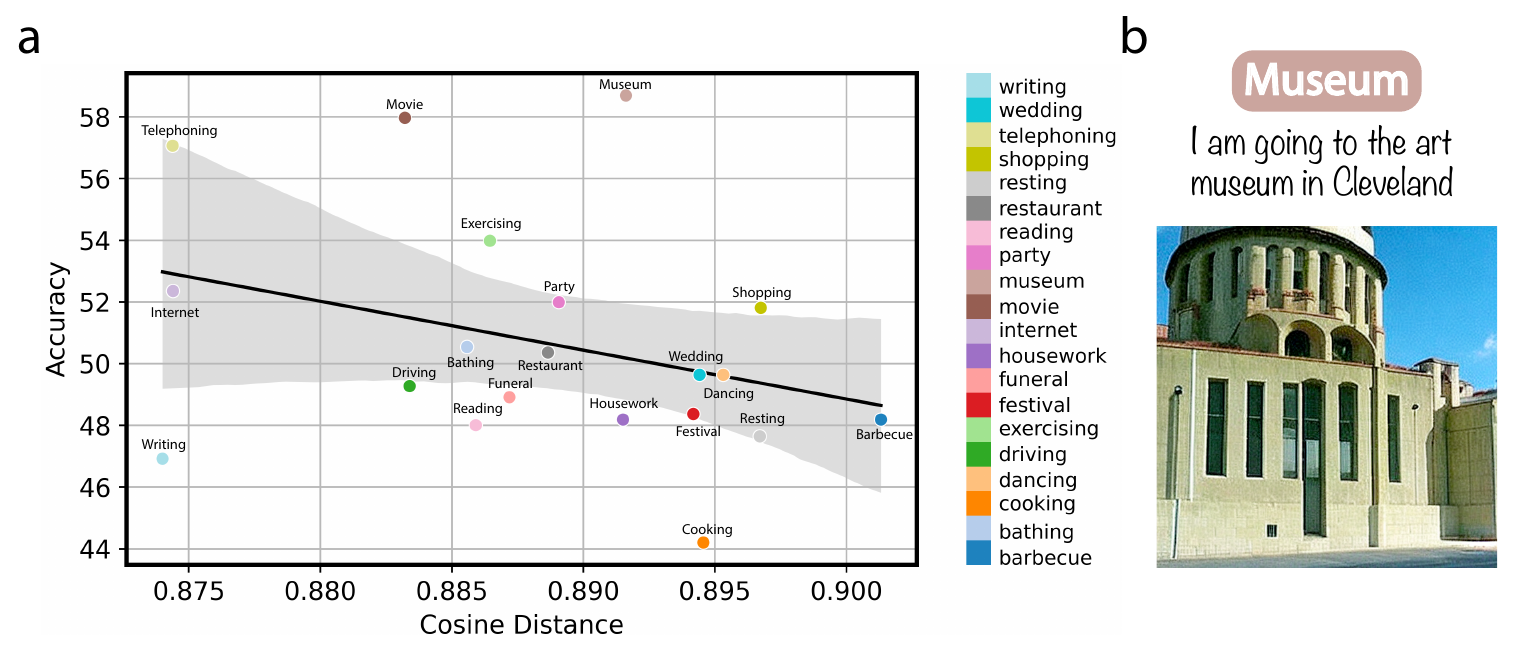}
    \caption{\blue{\textbf{Generalization to Zero-Shot Mental Imagery.}} a, \blue{WAVE performance on an independent dataset ($N=24$), where participants imagined scenarios based on verbal prompts.} This analysis measures the cosine distance of each scenario's features to those in the training dataset~(BOLD5000). The scatter plot shows the correlation of these cosine distances with decoding accuracy: scenarios that are more similar to the training dataset had more accurate predictions. b, Example of WAVE model zero-shot reconstructed images from imagination recording fMRI sessions. The text stimuli describing the museum scenario are presented above the example image, illustrating the model's capability to generate visual reconstructions based on described scenarios.}
    \label{fig:figure_5}
\end{figure}

To assess the WAVE model's capability in decoding semantic meanings from human brain activities, we conducted a zero-shot analysis using an imagination fMRI dataset. This analysis involved reconstructing images from 24 young adults' imaginations and comparing them with corresponding ground truth textual stimuli. We performed a two-way identification accuracy assessment across 480 paired images and texts, implementing a permutation test to validate the results. Our findings yielded an accuracy of 52\%, with a significant p-value of \textbf{0.0206}, indicating that our model effectively captures semantic meanings from brain activity associated with text-based imagination.

Additionally, we undertook a scenario-based scale analysis as illustrated in Figure~\ref{fig:figure_5}. Using a dataset in which participants provided verbal descriptions for scenarios that they later imagined during fMRI, we performed similar decoding of different scenarios using our model. The analysis revealed a correlation between the proximity of scenarios to the training dataset and the accuracy of decoding; specifically, closer cosine distances between scenarios and training data resulted in higher accuracy. For example, ``internet" or ``movie" scenarios that shared more semantic information with the training images were more easily decoded compared to ``barbecue" or ``resting" scenarios that were more semantically distinct. This scale analysis suggests that augmenting the training dataset could enhance our model's ability to decode imagined content by leveraging semantic brain activities, facilitated by the pre-trained whole-brain visual decoding capabilities of the WAVE model. See more Method details in Sec.~\ref{sec:zero-shot}.
\section{Discussion}
In this paper, we presented a novel approach to decode visual representations present throughout the brain by leveraging a pre-trained foundation model. Compared to the state-of-the-art approaches to decoding visual processing, we found that our model improved on predictive accuracy particularly when assessing the higher-level visual and semantic similarity between reconstructed and visual images. Saliency maps visualized regions most important to these predictions and identified predominantly regions in higher-order brain networks such as the frontoparietal and default mode networks, suggesting that higher-order processing was critical to this decoding. In line with this, masking the visual network still produced high (although reduced) levels of accuracy on the same visual decoding task. Ablating each network in turn showed that after the visual network, the dorsal attention and default mode networks contributed the most to predictions. This is in line with previous research: the dorsal attention network is adjacent to the visual network and performs higher level visual functions. The default mode network has a key role in semantic processing and ``making sense"~\cite{yeshurun2021default} of visual input, as shown by the presence of a retinotopic code in this network~\citep{steel2024retinotopic}. Post-hoc analyses comparing our model to others suggest that our model was better able to model the semantic information in images due to improved identification of key semantic features, sometimes with an associated loss of accuracy in lower-level visual similarity such as pixel correlation. \blue{Furthermore, our scaling analysis on an independent imagination dataset demonstrates the model's capacity for zero-shot generalization. We observed that decoding accuracy was modulated by semantic alignment: scenarios conceptually closer to the training distribution (e.g., ``internet'' or ``movie'') were reconstructed with higher fidelity than distinct concepts (e.g., ``barbecue''). This gradient in performance highlights that the model relies on shared semantic representations rather than memorized visual patterns. Crucially, these results validate the utility of foundation models for whole-brain analysis. Whole-brain fMRI is characterized by a massive feature space, high noise-to-signal ratio, and complex non-linear challenges where standard linear models or MLPs often struggle to converge. By leveraging pre-trained temporal and spatial priors, our approach effectively extracts signal from this high-dimensional noise, unlocking the rich semantic information encoded in non-visual networks like the dorsal attention and default mode networks, even when applied to datasets with very few individuals.}

In addition to our primary finding of improved performance over other state-of-the-art approaches, there are several important implications of our work for neuroscience. The ability to improve performance on visual processing using a foundation model trained and applied to the whole brain adds to recent literature highlighting the importance of non-visual regions to visual processing. For example, a recent study identified a retinotopic code in the default mode network that is hypothesized to provide a mapping between past experiences and incoming visual information~\citep{steel2024retinotopic}. This is in line with ideas from predictive coding~\citep{aitchison2017or} and active inference~\citep{friston2009reinforcement} theories of the brain that propose incoming visual information is continuously compared to an internal model of the world provided via top-down connections between visual and higher-order brain regions. The default mode network has been proposed to be critical in this ``sense-making"~\cite{yeshurun2021default}, in line with our finding that it contributed more than other non-visual networks to improved predictions. Related work suggests that visual images may be encoded in non-visual regions in terms of the actions they afford ~\citep{thornton2024neural}, in line with theories that consider the fundamental role of the brain to induce actions and predict their consequences as a means of increasing the survival of the individual~\citep{buzsaki2019brain}. Conceptually, our work builds on this by highlighting that higher-order brain networks, particuarly the default mode network, contribute critical information to visual decoding and allow for improved separation of visually similar images into distinct clusters based on higher-order visual and semantic information: for example, the separation of wild dogs from domestic dogs. 

Our post-hoc explainability analysis also provides some information about exactly which brain regions are contributing to decoding different categories of images. While it is difficult to interpret exactly what certain brain activity patterns mean for the information they represent without engaging in overly speculative reverse inference~\citep{poldrack2011inferring}, these data do add important interpretability to our findings in several ways. Breaking down these activation patterns by large-scale brain network, regions in the default mode network were important for all five clusters, in line with the role of this network in semantic processing~\citep{brandman2021surprising} and matching the results of our network-ablation anaylsis. The frontoparietal and dorsal attention networks were also important for predictions in all five clusters: these networks are involved in domain-general task-related processing and higher-order visual processing~\citep{marek2018frontoparietal,vossel2014dorsal} and this analysis suggests they contain information for decoding visual information in various domains. Some clusters did show unique ROIs: limbic regions were only involved in predictions of pets, potentially reflecting the particularly emotional nature of information processing of these images. Sensorimotor regions were most important for decoding the cluster representing wild animals, as shown in Fig.~\ref{fig:k-cluster}g, suggesting a possible additional engagement of areas involved in spatial awareness and movement perception~\citep{grosbras2012brain} when processing these images. This analysis further emphasizes the importance of whole-brain non-visual regions for visual processing and further suggests that certain images may engage different distributed neural processes depending on their contents.

\blue{We acknowledge several limitations in the current study. First, while our generative reconstructions achieve high semantic accuracy, we caution that diffusion models can introduce ``hallucinated'' high-frequency details based on their training priors. Therefore, the pixel-wise accuracy of our reconstructions should be interpreted as a synthesis of neural decoding and generative interpolation. Second, our claims regarding the "causal" contribution of the DMN are based on computational lesioning (masking); future work using transcranial magnetic stimulation (TMS) or patient studies would be required to establish true biological causality.}

The final important implication of this work is methodological. We believe that our approach allows for significant further advances in understanding the role of non-visual regions in visual processing by \blue{transferring learned priors from vast, non-specific fMRI repositories to specialized tasks.}
Moreover, \blue{this capability holds particular promise for advancing brain-behavior analyses, especially for complex traits that exhibit high inter-subject variability.} These models allow us to take advantage of the large amounts of non-specific fMRI data that now exist and may help us improve our predictions of these more complex brain-behavior relationships, beyond functional decoding and into more trait-level brain-behavior analyses. This has huge implications for the field, as these tasks have previously required the collection of either highly specific data or more general data in thousands of individuals~\citep{marek2022reproducible}. This has led to significant consternation in the field regarding the practicality of using fMRI to understand behavior and led some researchers to suggest that data for understanding brain-behavior relationships for complex traits should only be done by consortia~\citep{gratton2022brain}. Using similar principles to those proposed by other researchers~\citep{he2022meta}, \blue{brain foundation model captured universal neural activities can significantly boost predictive power in smaller datasets. Consequently, this approach has the potential to democratize robust brain-behavior research, allowing for precise modeling of complex cognitive traits without the prohibitive requirement of massive sample sizes.}
\section{Methods}
\label{sec:methods}

\subsection{Dataset}
\label{method:dataset}
\paragraph{BOLD5000}
For our study, we utilized the BOLD5000 dataset, a publicly available fMRI image collection~\citep{chang2019bold5000}. This dataset comprises fMRI recordings captured as subjects viewed a series of images. The BOLD5000 dataset contains 4,916 unique images, of which 2,000 are sourced from the COCO dataset~\citep{lin2014microsoft}, and 1,916 images are from ImageNet~\citep{deng2009imagenet}. In each trial, subjects were exposed to each image as a visual stimulus. Among these, 4,803 images were presented once per trial, while 113 images were shown multiple times, either three or four times across several trials, resulting in a total of 5,254 stimulus trials. The dataset documents responses from four subjects, three of whom completed all 5,254 recording trials. In our analysis, we focus on the data from all four subjects who completed the full or partial set of trials.

\blue{We selected BOLD5000 to capture the comprehensive dynamics of high-level visual experience. While widely used repositories like the Natural Scenes Dataset (NSD)~\cite{allen2022massive} prioritize high-throughput sampling via rapid presentation (3s stimulus, 1s gap), such rapid pacing effectively truncates the temporal window necessary for slow, top-down cognitive processes to fully unfold. On the other hand, BOLD5000 employs a slow event-related design with extended inter-stimulus intervals (9s gap). This extended duration allows for the complete emergence of delayed feedback signals and preserves the rich temporal dynamics required by our pretrained autoregressive fMRI encoder~\cite{thomas2022self}. This enables us to decode higher-order semantics from \textit{whole-brain} activity, capturing the full scope of the visual experience.}


We adopt the standard train/test split employed in existing BOLD5000 image reconstruction studies, as outlined in~\cite{chen2023seeing}. For the training set of each subject, we utilized trials with non-repeated image stimuli, encompassing a total of 4,803 samples. For the test set, we consolidated repeated image stimulus trials, resulting in 113 test samples. Furthermore, BOLD5000 provides labels for each image, which we incorporated as the text modality in our analysis.

\paragraph{fMRI Imagination Dataset}
Additionally, we employed a validation dataset from fMRI imagination recordings as per~\cite{anderson2020decoding}. This dataset includes 20 different scenarios such as resting, dancing, and museum visits. Subjects provided varied text descriptions for these scenarios. The fMRI data were recorded as subjects imagined themselves in the scenarios prompted by the text descriptions they had written. Each scenario was repeated five times during the fMRI sessions. In our zero-shot analysis, we averaged the fMRI signals from the repeated scenarios to create a single representative signal for each scenario. We engaged 24 young subjects for this study, resulting in a total of $24 \times 20 = 480$ imagination fMRI clips accompanied by text stimuli.

\paragraph{Upstream Datasets}
In this paper, we utilize a foundational fMRI model incorporating several large-scale fMRI datasets for pre-training~\citep{thomas2022self}. These include the Human Connectome Project~\citep{van2013wu} and multiple datasets available on OpenNeuro~\citep{markiewicz2021openneuro}, comprising a total of 11,980 fMRI runs from 1,726 individuals across 34 datasets. \blue{Crucially, the pre-training curriculum includes not only resting-state acquisitions but also a wide array of active tasks, such as passive visual perception (e.g., movie-watching), working memory recall, language processing, and motor coordination. This exposure to varied tasks allows the model to learn generalized spatiotemporal representations that transcend specific experimental paradigms.} Additionally, we employed a pre-trained image classifier to evaluate semantic accuracy metrics. This classifier was originally trained on ImageNet1K~\citep{russakovsky2015imagenet}, a dataset that includes approximately 1.2 million images distributed over 1,000 categories, each representing a distinct object or concept.

\paragraph{Preprocessing}
\label{sec:preproc}
Functional Magnetic Resonance Imaging~(fMRI) data are captured in a four-dimensional space, encompassing three spatial dimensions and one temporal dimension. The Blood-Oxygen-Level-Dependent~(BOLD) signals are organized in a sequence denoted as $S = \{V_t | t \geq 0\}$, where $t$ represents the Repetition Time~(TR). At each time point $t$, the volume $V_t$ is represented in a three-dimensional space as $V_t \in \R^{x\times y\times z}$. 

We processed all fMRI data using fMRIPrep, following the methodology outlined in~\cite{esteban2019fmriprep},  with adaptations to include surface preprocessing. Our preprocessing regimen aligns with the protocols described by~\cite{thomas2022self}employing the default settings of fMRIPrep with the exception of surface preprocessing. Subsequently, we performed a series of minimal additional processing steps on the derivatives from fMRIPrep. These steps included spatial smoothing of the fMRI sequences, detrending, high-pass filtering, and the removal of confounding factors.

\begin{figure}[t]
    \centering
   \includegraphics[width=\linewidth]{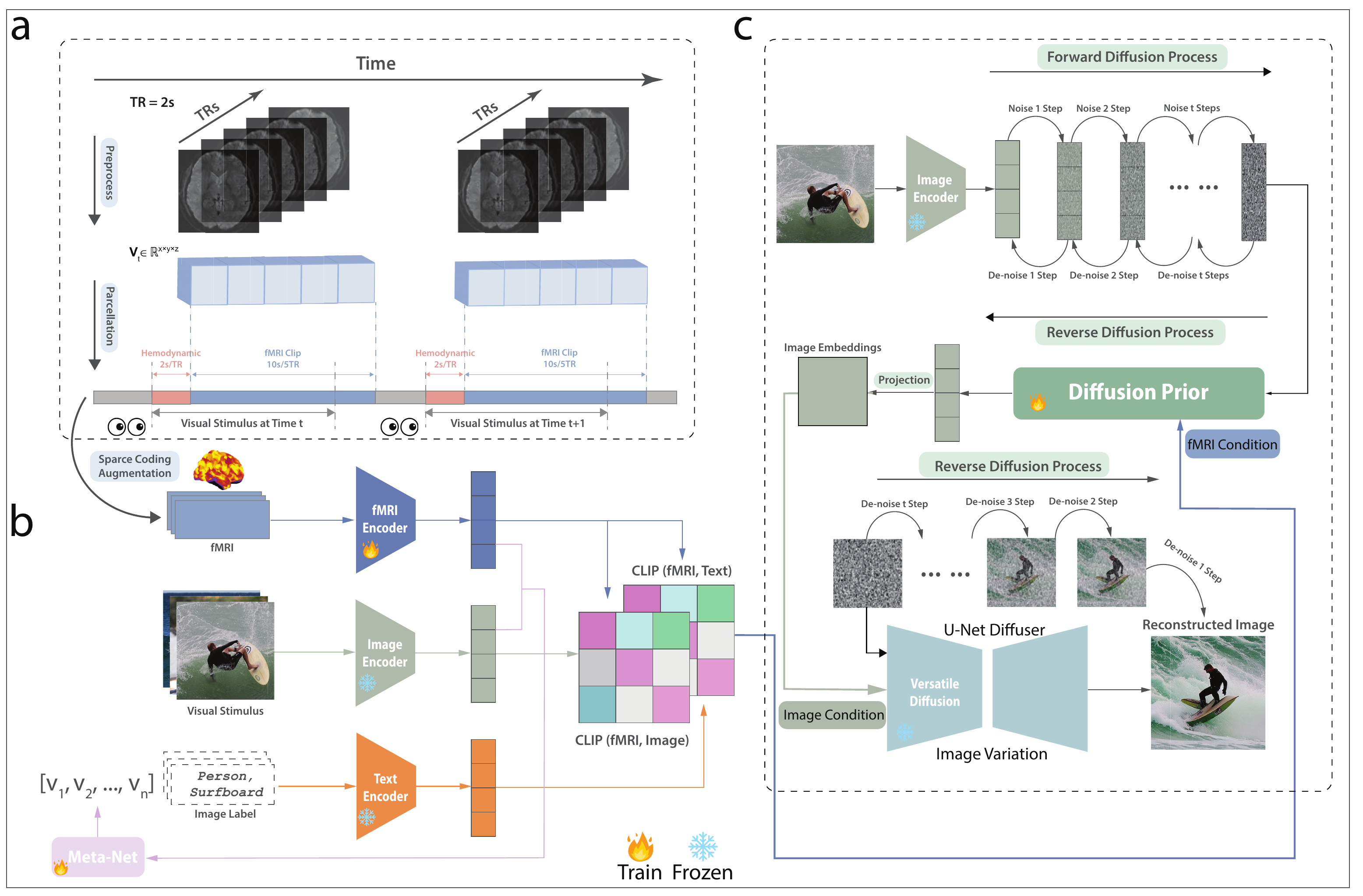}
   \vspace{-20pt}
    \caption{The depicted architecture illustrates the fMRI data processing and a two-part training approach for the model. a, fMRI preprocessing involving network parcellations and BOLD signal segmentation. b, Focus on contrastive learning where knowledge is distilled across three modalities: fMRI, text, and images. c,  Training of the diffusion model, which involves fine-tuning a specialized prior to converting fMRI latent representations into image latent variables. Icons of fire and snowflake denote modules that are active (train) and inactive (frozen) during the training phase, respectively.}
    \label{fig:bold_seg}
\end{figure}

After the preprocessing steps, we parcellated each preprocessed fMRI run using the DiFuMo atlas~\citep{dadi2020fine}, normalizing the individual network time courses to have a mean of zero and unit variance. We opted for the DiFuMo atlas with 1024 dimensions, effectively transforming each volume $V_t \in  \R^{x \times y \times z}$ into a vectorized form $X_t \in \R^{1024}$ as shown in Fig.~\ref{fig:bold_seg}. 

In the subsequent step of segmenting BOLD clips, we took into account the hemodynamic effects on BOLD signals\citep{pinsky2005functional}. To accommodate the delay introduced by these hemodynamic effects, we adjusted our segmentation to account for a one TR delay.
 In our dataset configuration, each segmented BOLD clip is characterized by five TRs, resulting in the final fMRI input for our model being $X \in \R ^{5 \times 1024}$, where $X$ represents our input data.

\subsection{WAVE Model}
Our model employs state-of-the-art deep learning~\citep{lecun2015deep,goodfellow2016deep} and foundation model~\cite{bommasani2021opportunities} techniques, structured into two main phases: contrastive learning~\citep{radford2021learning} and diffusion model training~\citep{rombach2022high}. \blue{Unlike approaches that rely on hybrid training schemes~\cite{scotti2024reconstructing}, our framework leverages a progressive learning framework for better fMRI foundation model adaption~\citep{rusu2016progressive}.} We initiate with contrastive learning to refine initial insights, followed by the diffusion model phase for enhanced training. Decoding fMRI data poses significant challenges due to its complexity. To address this, we incorporate a pre-trained large fMRI foundation model\citep{thomas2022self}. In the contrastive learning stage, we integrate image labels as an additional modality, recognizing that label data alone may not provide comprehensive information. To bridge this gap, we employ prompt learning to enrich the text encoder with supplementary image and fMRI details. For the diffusion model training, we introduce a novel approach by employing a diffusion prior\citep{ramesh2022hierarchical,scotti2024reconstructing} to transform fMRI latent representations into image latent representations. This process is completed with a Pre-trained Versatile Diffusion Image Variation Decoder\citep{xu2023versatile}, which translates these image latent representations back into visual images. This methodology not only leverages advanced machine learning techniques but also introduces an innovative integration of fMRI data into the image generation process, enhancing the model's ability to interpret and visualize complex neuroimaging datasets.

\paragraph{fMRI Foundation Model}
\blue{To leverage large-scale neural representations, we utilize the fMRI foundation model developed by Thomas et al.~\cite{thomas2022self}, which employs a decoder-only Transformer architecture (GPT-2)~\cite{radford2019language}.} As specified in
Section~\ref{sec:preproc}, the input data $X$, is conformed to the preprocessed structure $X \in \R^{t \times 1024}$, where t is the number of TRs and 1024 denotes the parcellated brain networks. The fMRI Foundation model employs self-supervised learning~(SSL) on extensive fMRI datasets, including the Human Connectome Project~\citep{van2013wu} and datasets available through OpenNeuro~\citep{markiewicz2021openneuro}. It assimilates both the sequential temporal fMRI information and TR embeddings, which function similarly to position embeddings~\citep{vaswani2017attention} to indicate the temporal intervals between fMRI sequences. The self-supervised learning task is to predict the subsequent timepoint sequence given an input sequence $X_t$; thus, the model predicts $X_{t+1}$.

In this study, we processed an fMRI run, retaining the spatial and temporal features to form the input $X$, where $X \in \R^{5 \times 1024}$. Each fMRI segment has 5 TRs. The attention is on the embedded fMRI sequence. Based on the SSL paradigm, we took the predicted masked token, which is the output $O_t \in \R^{768}$ where $t = 6$. The output $O_6$ was taken as our encoded fMRI latent features. \blue{Please refer to Appendix~\ref{app:method} for more details.}

\paragraph{Contrastive Learning}
In this study, we employ multi-modal contrastive learning, capitalizing on the integration of diverse data forms to enrich the feature representation of fMRI scans in a latent space. Our approach is distinct in that it encompasses not only visual data but also textual information through the inclusion of image labels. This incorporation aims to augment the semantic understanding within our model. To facilitate this, we utilize a pre-trained Vision Transformer~(ViT)~\cite{dosovitskiy2020image} based CLIP model~\cite{radford2021learning}, maintaining the integrity of its parameters by freezing them throughout the training phase.

Our implementation introduces CLIP loss within the contrastive learning framework, addressing three different modalities: fMRI-image and fMRI-text. Let the normalized projections of modality as $z$ and a temperature parameter as $\mathcal{T}$, the CLIP Loss $\mathcal{L_{CLIP}}$ of two modalities $z_1, z_2$ in batch $B$ is shown in Eq.~\ref{form:clip_loss}. The computation of contrastive Loss $\mathcal{L_{C}}$~(Eq.~\eqref{form:contrastive_loss}) involves aggregating the losses from fMRI to image and fMRI to text comparisons, followed by averaging the sum. The training regimen proceeded over 50,000 steps utilizing an Adam optimizer, selecting the model state at the final step for subsequent training with a diffusion model.

\begin{equation}
\small
    \mathcal{L_\text{CLIP}}(z_1, z_2) = -\frac{1}{2|B|} \sum_{i=1}^{|B|} \left[ \log \frac{\exp({z}_{\text{1}_i} \cdot {z}_{\text{2}_i} / \tau)}{\sum_{j=1}^{|B|} \exp({z}_{\text{1}_i} \cdot {z}_{\text{2}_j} / \tau)} + \log \frac{\exp({z}_{\text{2}_i} \cdot {z}_{\text{1}_i} / \tau)}{\sum_{j=1}^{|B|} \exp({z}_{\text{2}_i} \cdot {z}_{\text{1}_j} / \tau)} \right],
    \label{form:clip_loss}
\end{equation}

\begin{equation}
    \mathcal{L_{C}} = (\mathcal{L_\text{CLIP}}({z}_{\text{fMRI}}, z_{\text{image}}) + \mathcal{L_\text{CLIP}}({z}_{\text{fMRI}}, z_{\text{text}})) / 2.
    \label{form:contrastive_loss}
\end{equation}

Furthermore, we assess the efficacy of our training process through contrastive learning accuracy metrics. \blue{Detailed implementation of these methodologies and the results are provided in~\ref{app:method}.}

\paragraph{Prompt Learning}
To enhance our contrastive learning framework, we were inspired by prompt learning~\citep{zhou2022conditional,khattak2023maple,yang2024MMMA} in the AI field, which is particularly effective when text labels serve as one modality in contrastive learning. Text labels, while compact, inherently contain less information than images. Prompt learning, a technique tailored for text encoders, enriches this modality by generating context that extracts more nuanced information from the text. Considering the fMRI, image, and text share the same information. We introduced a lightweight architecture called Meta-Net designed to extract features from fMRI~$\mathcal{H_{\textit{fMRI}}}$ and image~$\mathcal{H_{\textit{image}}}$ latent space, $\mathcal{H} \in \R^{768}$, and $\text{features}~\mathcal{F} \in R^{1536}$ and output prompt learnable vectors $V \in \R^{8}$ for the text encoder. The vectors will be concatenated with tokenized text together input to the CLIP text encoder. Unlike CLIP text and image encoders, the gradients of the Meta-Net were not frozen during the contrastive training, where the Meta-Net was continuously trained and learnable. 

\paragraph{Diffusion Model}
In this study, we introduce a novel approach to fMRI decoding by proposing a diffusion prior integrated within a flexible diffusion model during the training phase. We design the diffusion prior with adjustable parameters, ensuring these are the only elements subject to learning while training the diffusion part. Distinctly, the training of the diffusion model progresses independently from the contrastive learning process. This method leverages latent fMRI features~$\mathcal{H_{\textit{fMRI}}}$, extracted by an fMRI encoder, which has been refined via contrastive learning across three modalities. To streamline our model, we eliminated the text and image encoders, retaining solely the fMRI encoder that produces latent~$\mathcal{H_{\textit{fMRI}}}$ in $\in \R^{768}$. 
The foundation of our diffusion prior is based on UNet~\cite{ronneberger2015u}, which utilizes the fMRI latent space as a conditional input. The primary function of this prior is to transform the fMRI latent into image latent space suitable for \blue{generative reconstruction}. During the training, we incorporate a singular linear layer with weights $W$ to adjust the dimensions of the fMRI latent, effectively transforming them as~$\mathcal{H_{\textit{fMRI}_{\text{T}}}}$ = $W \times \mathcal{H_{\textit{fMRI}}}$, with the transformed latent space having dimensions $\mathcal{H_{\textit{fMRI}_{\text{T}}}} \in \R^{257 \times 768}$. 
Our training process employs an $\ell_2$ loss $\mathcal{L_{D}}$ to measure the discrepancy between the ground-truth image latent~$\mathcal{H_{\textit{image}}}$, encoded by a pre-trained CLIP image encoder, and the image latent~$\mathcal{H_{\textit{image}_{\text{D}}}}$ denoised by our diffusion prior from the fMRI latent~$\mathcal{H_{\textit{fMRI}_{\text{T}}}}$. We enhance the robustness of our model by applying intensive data augmentation techniques to the input fMRI data, including sequence permutation, normalization, and sparse coding. Furthermore, as a regularization strategy, we adjust the impact of the loss by applying a loss multiplier~$\alpha$ of 0.3 as $\alpha \mathcal{L_{D}}$, optimizing the model's performance and generalization capability. 

After the prior training, we also integrate the versatile diffusion model as the decoder to decode the converted image latent from fMRI by the diffusion prior to the actual image. We used 20 steps of diffusion to reconstruct the final image. Additionally, we also apply an image selection technique using our encoder to calculate the CLIP similarity matrix between the reconstructed image and the input fMRI. See details in Appendix. We used our trained fMRI encoder to encode the input fMRI and a pre-trained CLIP image encoder to encode the reconstructed image. We select the best one. In this study, we constructed 2 images at the same time. 

\paragraph{Metrics}\label{method:metrics} Following previous works~\citep{chen2023seeing,gaziv2022self} measurement for reconstructed images, we used the 50-way-top-1 accuracy classification task to evaluate the semantic correctness of the reconstructed images. As the same process as \cite{chen2023seeing}, we used a pre-trained ~\citep{dosovitskiy2020image,paszke2017automatic} ImageNet1K classifier. Both ground-truth and reconstructed images were input to the classifier, and we checked if the top-1 classification of the generated image matched the ground-truth image classification. This metric straightforwardly examines the reconstructed image's semantic meaning through classification accuracy. This metric is the most critical since the model is aiming to recover the semantic meaning of the images in this work.

In addition to evaluating reconstructed image semantic meaning, we followed previous works by adopting both low-level and high-level metrics as supplementary evaluation for reconstructed image detail features. In low-level metrics, we used pixel-level correlation(PixCorr) to evaluate the reconstructed images and ground-truth images. We used two-way identification to compare the second and fifth layers of AlexNet, and the last pooling layer of InceptionV3, and the output layer of the CLIP-Vision model. EffNet-B~(Eff) and SwAV-ResNet50~(SwAV) are used to calculate the distance correlation. Please see~\cite{ozcelik2023natural} for more details.

\subsection{Whole-Brain Visual Experience Analysis}
We \blue{developed} a self-supervised \blue{analytical framework to map distinct visual semantic categories to their corresponding whole-brain representations.} This method \blue{examine specific brain region activities covary with distinct clusters of visual images. By partitioning the stimulus set into five semantically coherent clusters, this analysis} enhances our understanding of the nuanced relationships between visual stimuli and brain responses.

\paragraph{\blue{Semantic} Clustering}
\blue{To investigate the neural correlates of distinct visual concepts, we employed an unsupervised clustering approach on the stimulus set. Since direct clustering of raw pixel values often captures low-level statistics rather than semantic content, we first projected all visual stimuli into a high-dimensional semantic space using the pre-trained CLIP image encoder. We then applied the K-Means algorithm to these latent embeddings to partition the dataset into five distinct semantic clusters (Fig.~\ref{fig:k-cluster}). To interpret the thematic content of each cluster, we aggregated the object labels associated with the images in each group and visualized them as word clouds, where font size corresponds to label frequency. Finally, to derive cohesive category definitions, we utilized a Large Language Model (GPT-4) to synthesize the dominant themes from the label distributions. The resulting category definitions are }``Pets", ``Outdoor Activities", ``Wild Animals", ``Interiors", and ``Indoor Objects." \blue{We also manually verified to ensuring they accurately reflected the visual contents of each cluster.}

\paragraph{Gradient-based Saliency Map}
To assess the model's focus on specific neural networks, we utilized Grad-CAM~\citep{selvaraju2020grad}, a gradient-based technique. This method involves tracking the gradients from the target output back through the network layers to the originating fMRI input data. This process helps in pinpointing which areas of the input are most influential in the model's predictions, where the gradient values are high. In our cluster analysis, the target outputs are the clusters. We created gradient-based saliency map for each cluster. As shown in Fig.~\ref{fig:k-cluster}, we averaged the gradient values from four subjects in the dataset, and we took the top 20 regions for visualization.

\paragraph{Decoding Visual Experiences Beyond the Visual Cortex}
During the training and reconstruction phases, we incorporated a visual region mask into the fMRI input data. This mask, aligned with the DiFuMo atlas~\citep{dadi2020fine}, sets the values within the visual regions to zero by following the network mapping scheme to the Yeo 7 networks~\cite{yeo2011organization}. Specifically, we applied the mask to all input fMRI data associated with the visual network as defined in the Yeo 7 classification. This masking procedure was consistently maintained throughout the training and image reconstruction processes. The images reconstructed from the fMRI data with the visual networks masked are displayed in Fig.~\ref{fig:vis_mask_examples}. The ablation network ablation of other six networks are the same process described as visual network above.

\subsection{Zero-Shot Human Imagination Decoding}
\label{sec:zero-shot}
We applied our pre-trained WAVE model in a universal setting to the BOLD5000 dataset, aiming for zero-shot reconstruction of images from fMRI imagination recordings. The quality of reconstructed images was evaluated using two-way identification metrics, comparing them to the ground-truth text stimuli. The decoding capability of the model is closely linked to the scale and distribution of the training dataset. To further explore the potential of imagination image decoding, we conducted a scale analysis, contrasting imagination scenarios with training images.

\paragraph{Two-Way-Identification Accuracy}
To assess the quality of the reconstructed images against the ground truth texts, we utilized the Two-Way-Identification Accuracy metric. This involved using a pre-trained CLIP model to encode \(n\) reconstructed images and \(n\) ground-truth texts into image embeddings \(\mathcal{H_{\textit{image}}}\) and text embeddings \(\mathcal{H_{\textit{text}}}\), respectively. We computed the Pearson product-moment correlation coefficients~\cite{harris2020array}, resulting in an image-text coefficient matrix \(M \in \mathbb{R}^{n \times n}\). The accuracy for each image was determined by the ratio of row elements with values less than their diagonal counterparts, divided by \(n-1\). The overall two-way identification accuracy is the average of these individual accuracies.

We recruited 24 young \blue{adults~(age under 30)}, each \blue{participant was} presented with 20 scenarios accompanied by text descriptions. Utilizing the aforementioned metric, we aggregated the data into a comprehensive coefficient matrix \(M_{\textit{all}} \in \mathbb{R}^{480 \times 480}\), and performed a permutation test 5,000 times by randomizing the sequence of image embeddings.

\paragraph{Scale Analysis}
In our scale analysis, we constructed a scenario-based image-text coefficient matrix \(M_{\textit{sc}} \in \mathbb{R}^{24 \times 24}\) to quantify the scenario-based accuracies. To further explore the interplay between each scenario and the images from the WAVE training set, we computed the cosine distances, resulting in a cosine distance matrix \(M_{\textit{d}} \in \mathbb{R}^{24 \times 4803}\). This matrix represents the distances between the textual descriptions of each scenario and the 4,803 training images. To synthesize these relationships into actionable insights, the mean cosine distance was calculated for each scenario using the formula:

\begin{equation}
\mathcal{D_{\text{mc}}} = \frac{1}{|U| \times |V|} \sum_{u \in U} \sum_{v \in V} \left(1 - \frac{u \cdot v}{\|u\|_2 \|v\|_2}\right).
\label{form:mean_consine_distance}
\end{equation}

In Eq.~\ref{form:mean_consine_distance}, \(U\) and \(V\) are the sets of vector representations for the scenario texts and image features respectively, both with dimensionality \(768\), where \(|U| = 24\) and \(|V| = 4803\). The formula calculates the average cosine distance across all pairs formed between a vector \(u\) in \(U\) and a vector \(v\) in \(V\), with \(u \cdot v\) denoting the dot product and \(\|u\|_2\), \(\|v\|_2\) their respective Euclidean norms. The results from this computation are depicted in Fig.~\ref{fig:figure_5}, illustrating the nuanced relationship between textual scenarios and visual training data.

\section*{Acknowledgment}
This research is supported by the HAI-Google Research Award, UST, HAI Hoffman-Yee Award, and National Institutes of Health grants NIH U24AG072701 and R01AG089169. We also greatly thank Yurong Liu from New York University for providing figure suggestions and proofreading.

\section*{Code and data availability}
To ensure the reproducibility of our analyses, we have made our code and processed data publicly available. These resources can be accessed via our GitHub repository at \url{https://github.com/PPWangyc/WAVE}. See more details at our project website:~\url{https://wave-brain.github.io}.

Our processed datasets and pre-trained models are available in our~\href{https://huggingface.co/collections/PPWangyc/wave-66808dc22b1f40e838c43b97}{Hugging Face collection}. Specifically, the processed Whole-Brain BOLD5000 dataset is available at 
\url{https://huggingface.co/datasets/PPWangyc/WAVE-BOLD5000}
, where you can also find our pre-trained models at \url{https://huggingface.co/PPWangyc/WAVE-models}. Additionally, accessing the extra validation dataset of fMRI imagination requires IRB approval and a Data Use Agreement (DUA). For more information or to request access, please contact us directly.





\bibliographystyle{elsarticle-num}
\bibliography{refs}

\appendix
\clearpage
\section{Appendix / Supplemental Material}
    \subsection{Individual Decoding Result}
    \paragraph{Whole Brain Decoding.}
    In our main paper, we present two experiment settings, the individual model for decoding and the universal model. To explore our WAVE model decoding ability for individual subjects. We provide each subject metric results here. As shown in the Table~\ref{tab:table_3}. We showed all four subjects in the BOLD5000 dataset, and each subject ID is listed in the first column, and the details metrics results are followed in each row. WAVE has great advances in decoding semantic meaning from the fMRI and the results in High-Level and Semantic Top-1 Accuracy also prove this point. \blue{these results demonstrate that our model maintains high performance stability across all subjects (CSI1--CSI4). The inter-subject variance is notably low, particularly in high-level semantic metrics (e.g., CLIP and Top-1 Accuracy), where the model consistently outperforms baselines.}
    \begin{table}[h]
    \caption{\textbf{Individual Subject Reconstructed image evaluation metrics.} We extend our averaged across subject result table in manuscript to individual result. All the metrics are the same setting as the manuscript.}
    \begin{adjustbox}{width=\columnwidth}
    \begin{tabular}{c|c|cccc|cccc|c}
    \toprule
    \multirow{2}{*}{\textbf{Subject}} &
      \multirow{2}{*}{\textbf{Model}} &
      \multicolumn{4}{c|}{\textbf{Low-Level}} &
      \multicolumn{4}{c|}{\textbf{High-Level}} &
      \textbf{Semantic} \\ 
     &          & PixCorr$\uparrow$ & SSIM$\uparrow$          & AlexNet(2)$\uparrow$ & AlexNet(5)$\uparrow$ & Incep$\uparrow$  & CLIP$\uparrow$   & Eff$\downarrow$  & SwAV$\downarrow$ & Top-1 Acc$\uparrow$ \\ \hline
    \multirow{3}{*}{CSI1} &
      WAVE~(Ours) &
      0.078 &
      0.230 &
      76.29\% &
      83.00\% &
      \textbf{75.20\%} &
      \textbf{86.37\%} &
      \textbf{0.859} &
      0.539 &
      \textbf{33.28\%} \\
     & MindEye~\citep{scotti2024reconstructing}  & 0.079    & 0.226          & \textbf{76.90\%}     & \textbf{84.06\%}     & 67.67\%   & 67.67\% & 0.895 & 0.566 & 17.78\%   \\
     & Mind-Vis~\citep{chen2023seeing} & \textbf{0.095}    & \textbf{0.313} & 73.81\%     & 82.18\%     & 74.02\% & 70.07\% & 0.877 & \textbf{0.514} & 23.64\%   \\ \hline

     \multirow{3}{*}{CSI2} &
      WAVE~(Ours) &
      0.061 &
      0.167 &
      \textbf{72.38\%} &
      \textbf{82.74\%} &
      \textbf{72.06\%} &
      \textbf{82.71\%} &
      \textbf{0.890} &
      \textbf{0.548} &
      \textbf{28.53\%} \\
     & MindEye~\citep{scotti2024reconstructing}  & 0.046    & 0.218          & 67.14\%     & 72.18\%     & 60.99\%   & 68.09\% & 0.940 & 0.616 & 11.01\%   \\
     & Mind-Vis~\citep{chen2023seeing} & \textbf{0.076}    & \textbf{0.320} & 69.90\%     & 77.09\%     & 68.39\% & 70.07\% & 0.906 & 0.546 & 16.04\%   \\ \hline

     \multirow{3}{*}{CSI3} &
      WAVE~(Ours) &
      0.063 &
      0.208 &
      \textbf{70.14\%} &
      \textbf{78.67\%} &
      65.40\% &
      \textbf{80.04\%} &
      \textbf{0.898} &
      \textbf{0.550} &
      \textbf{23.00\%} \\
     & MindEye~\citep{scotti2024reconstructing}  & 0.031    & 0.218          & 69.95\%     & 73.32\%     & 64.68\%   & 67.53\% & 0.934 & 0.611 & 14.32\%   \\
     & Mind-Vis~\citep{chen2023seeing} & \textbf{0.085}    & \textbf{0.316} & 68.66\%     & 74.98\%     & \textbf{66.20\%} & 70.07\% & 0.912 & 0.553 & 16.40\%   \\ \hline

     \multirow{3}{*}{CSI4} &
      WAVE~(Ours) &
      0.045 &
      0.193 &
      67.60\% &
      74.38\% &
      62.38\% &
      \textbf{77.04\%} &
      0.922 &
      \textbf{0.572} &
      16.15\% \\
     & MindEye~\citep{scotti2024reconstructing}  & \textbf{0.068}    & 0.207          & \textbf{70.49\%}     & \textbf{79.69\%}     & \textbf{71.99\%}   & 76.64\% & \textbf{0.895} & 0.574 & \textbf{24.27\%}   \\
     & Mind-Vis~\citep{chen2023seeing} & 0.043    & \textbf{0.286} & 57.33\%     & 56.73\%     & 59.08\% & 58.15\% & 0.961 & 0.610 & 14.55\%   \\ 
     \bottomrule
    \end{tabular}
    \label{tab:table_3}
    \end{adjustbox}

\end{table}
    
    \paragraph{\blue{Decoding from Non-Visual Networks.}}Besides presenting the decoding results from whole brain wise data (WAVE) and visual cortex data (MindEye \& Mind-Vis), we also explored decoding information from brain regions that do not belong to the visual networks. By applying the mask to the input fMRI data, WAVE can still successfully decode the natural image from fMRI across all subjects. In addition to the averaged subject results in the manuscript, we also provide the individual visual networks mask results in Table~\ref{tab:table_4}. Our individual visual mask result gives new insights into neuroscience to understand various brain regions' functions not only at visual perception level, but in the cognition. \blue{These findings provide critical neuroscientific insights, suggesting that visual experience is not confined to perceptual areas but is deeply integrated into higher-order cognitive functions. As expected, the quantitative metrics for this masked condition were slightly lower than the unmasked whole-brain results reported in Table~\ref{tab:table_3}.}
    \begin{table}[h]
    \caption{\textbf{Individual Subject visual network masked decoding results.} The brain regions belonging to the visual network according to Yeo 7 Atlas were masked in the fMRI input. The base model is WAVE.}
    \begin{adjustbox}{width=\columnwidth}
    \begin{tabular}{c|cccc|cccc|c}
    \toprule
    \multirow{2}{*}{\textbf{Subject}} &
      \multicolumn{4}{c|}{\textbf{Low-Level}} &
      \multicolumn{4}{c|}{\textbf{High-Level}} &
      \textbf{Semantic} \\ 
              & PixCorr$\uparrow$ & SSIM$\uparrow$          & AlexNet(2)$\uparrow$ & AlexNet(5)$\uparrow$ & Incep$\uparrow$  & CLIP$\uparrow$   & Eff$\downarrow$  & SwAV$\downarrow$ & Top-1 Acc$\uparrow$ \\ \hline
    \multirow{1}{*}{CSI1} &
      0.059 &
      0.191 &
      67.75\% &
      78.52\% &
      70.12\% &
      82.35\% &
      0.881 &
      0.558 &
      25.95\% \\
      
    \multirow{1}{*}{CSI2}&
      0.053 &
      0.185 &
      67.31\% &
      75.43\% &
      67.42\% &
      78.58\% &
      0.911 &
      0.591 &
      17.40\% \\

     \multirow{1}{*}{CSI3} &
      0.033 &
      0.176 &
      65.41\% &
      75.43\% &
      62.49\% &
      75.73\% &
      0.920 &
      0.600 &
      18.78\% \\

     \multirow{1}{*}{CSI4} &
      0.041 &
      0.193 &
      66.10\% &
      71.11\% &
      62.62\% &
      75.63\% &
      0.925 &
      0.603 &
      13.79\% \\
        \bottomrule
    \end{tabular}
    \label{tab:table_4}
    \end{adjustbox}

\end{table}

    \paragraph{Ablation Studies.}
        We conducted ablation studies to evaluate the performance of different encoder and decoder configurations in diffusion-based image reconstruction. Two types of encoders were used: the pre-trained fMRI Foundation model and the train-from-scratch MLP Backbone. For the decoders, we explored two diffusion models: Versatile Diffusion and Stable Diffusion. The ablation results, presented in Table~\ref{tab:decoder_ablation}, indicate that the fMRI Model, which employs a pretrained whole-brain fMRI encoder~\cite{thomas2022self}, demonstrates superior performance compared to the MLP Backbone. Additionally, Versatile Diffusion consistently outperformed Stable Diffusion across evaluated metrics.
\begin{table}[h]
\caption{\textbf{Encoder and Decoder Ablation Studies on Subject CSI3.} This table details our ablation studies comparing the Versatile and Stable diffusion models. Different encoder configurations were tested: the region encoder utilizes whole brain data, while the voxel encoder targets the visual cortex fMRI specifically. The fMRI Model refers to the pre-trained fMRI Foundation model, and the MLP Backbone is a custom-built residual MLP model. We assessed the models based on semantic accuracy and Fréchet Inception Distance (FID), with lower FID values indicating better image reconstruction quality.}
\centering
\begin{adjustbox}{width=\columnwidth}
\begin{tabular}{llccccccc}
\toprule
\multicolumn{2}{c}{\textbf{fMRI Encoder}} & \multirow{2}{*}{\textbf{Decoder}} & \multirow{2}{*}{\textbf{Acc(\%)}} & \multirow{2}{*}{\textbf{Min(\%)}} & \multirow{2}{*}{\textbf{Max(\%)}} & \multirow{2}{*}{\textbf{FID}} & \multirow{2}{*}{\textbf{Min}} & \multirow{2}{*}{\textbf{Max}} \\ 
\cmidrule(lr){1-2}
\textbf{Region Enc} & \textbf{Voxel Enc} & & & & & & & \\ 
\midrule
fMRI Model~(\cite{thomas2022self}) & w/o & VD & \textbf{22.13 ± 1.48} & \textbf{19.62} & \textbf{23.41} & \textbf{2.45 ± 0.71} & \textbf{1.68} & \textbf{3.55} \\ 
fMRI Model~(\cite{thomas2022self}) & w/o & SD & 18.81 ± 1.17 & 17.87 & 20.46 & 3.70 ± 1.48 & 2.37 & 5.77 \\ 
MLP Backbone & w/o & VD & 18.79 ± 1.72 & 16.71 & 20.92 & 4.81 ± 0.95 & 3.53 & 5.80 \\ 
MLP Backbone & w/o & SD & 13.03 ± 0.95 & 12.17 & 14.35 & 6.69 ± 1.55 & 4.94 & 8.70 \\ 
w/o & MLP Backbone & VD & 12.91 ± 1.12 & 11.41 & 14.09 & 3.65 ± 0.63 & 3.09 & 4.53 \\ 
w/o & MLP Backbone & SD & 10.41 ± 1.00 & 9.01 & 11.23 & 6.62 ± 1.36 & 4.87 & 8.20 \\ 
\bottomrule
\end{tabular}
\end{adjustbox}
\label{tab:decoder_ablation}
\end{table}

        In addition to exploring various encoder and decoder configurations, our research extended to analyzing the impact of decoder loss functions on Subject CSI3. Our findings indicate that utilizing diffusion model loss directly on image pixel values, rather than on embeddings, enhances performance in low-level pixel-related metrics and Fréchet Inception Distance (FID). As demonstrated in Table~\ref{tab:loss_ablation}, Mind-Vis~\cite{chen2023seeing} has better low-level performance. 
\begin{table}[h]
\caption{\textbf{Decoding loss ablation on Subject CSI3.} This table compares the performance of different models based on the type of loss function used, whether applied directly to the image pixels or to the latent embedding space. WAVE and MindEye utilize an embedding-based loss, whereas Mind-Vis employs a pixel-based loss with a Latent Diffusion Model (LDM) as the decoder.}
\centering
\begin{adjustbox}{width=\columnwidth}
\begin{tabular}{l|cccccccc}
\toprule
\multirow{2}{*}{\textbf{Model}} & \multirow{2}{*}{\textbf{Loss}} & \multirow{2}{*}{\textbf{Decoder}} & \multirow{2}{*}{\textbf{Acc(\%)}} & \multirow{2}{*}{\textbf{Min(\%)}} & \multirow{2}{*}{\textbf{Max(\%)}} & \multirow{2}{*}{\textbf{FID}} & \multirow{2}{*}{\textbf{Min}} & \multirow{2}{*}{\textbf{Max}} \\
& & & & & & & & \\
\midrule
WAVE & Embed & VD & \textbf{22.13 ± 1.48} & \textbf{19.62} & \textbf{23.41} & 2.45 ± 0.71 & 1.68 & 3.55 \\
MindEye~(\cite{scotti2024reconstructing}) & Embed & VD & 13.45 ± 0.13 & 12.89 & 14.97 & 2.91 ± 0.39 & 2.53 & 3.30 \\
Mind-Vis~(\cite{chen2023seeing}) & Pixel & LDM & 14.83 ± 2.25 & 12.59 & 19.08 & \textbf{1.64 ± 0.19} & \textbf{1.40} & \textbf{1.98} \\
\bottomrule
\end{tabular}
\end{adjustbox}
\label{tab:loss_ablation}
\end{table}

        We also conduct research to show multi-modal fMRI-Visual-Language learning with prompt Meta-Net effect in this decoding task. As illustrated in Table~\ref{tab:contrast_ablation}, incorporating multiple modalities through contrastive learning enhances the model's ability to encode representations, thereby improving the classification accuracy for Brain (fMRI) and Vision (Visual Stimulus) using the CLIP model. The introduction of prompt-based learning further aids in balancing the training across multiple modalities, enhancing both Brain-Vision and Brain-Language accuracy.
\begin{table}[h]
\caption{\textbf{Ablation Study for the Contrastive Learning Method on Subject CSI3.} This table presents the ablation results comparing the Brain-Vision contrastive learning model, which involves only visual stimulus and fMRI data, with models that include Brain-Vision-Language triple modality contrastive learning and Meta-Net Prompt triple modality learning. The evaluation metrics include Top-1 classification accuracy for CLIP, along with Brain to Vision Accuracy and Brain to Language Accuracy.}
\begin{adjustbox}{width=\columnwidth}
\begin{tabular}{l|cc}
\toprule
\textbf{Model} & \textbf{Brain-Vision Accuracy~(\%)} & \textbf{Brain-Language Accuracy~(\%)}                 
\\ \hline
Vision Only & 43.96\%               & 20.0\%                                        \\ \hline
Vision + Language w/o Prompt & 49.44\%                  & 34.81\%           \\ \hline
Vision + Language w/ Prompt~(WAVE)  & \textbf{53.98\%}                  & \textbf{36.28\%}            \\ 
\bottomrule
\end{tabular}
\end{adjustbox}
\label{tab:contrast_ablation}
\end{table}

\begin{figure}[]
    \centering
    \includegraphics[width=0.8\linewidth]{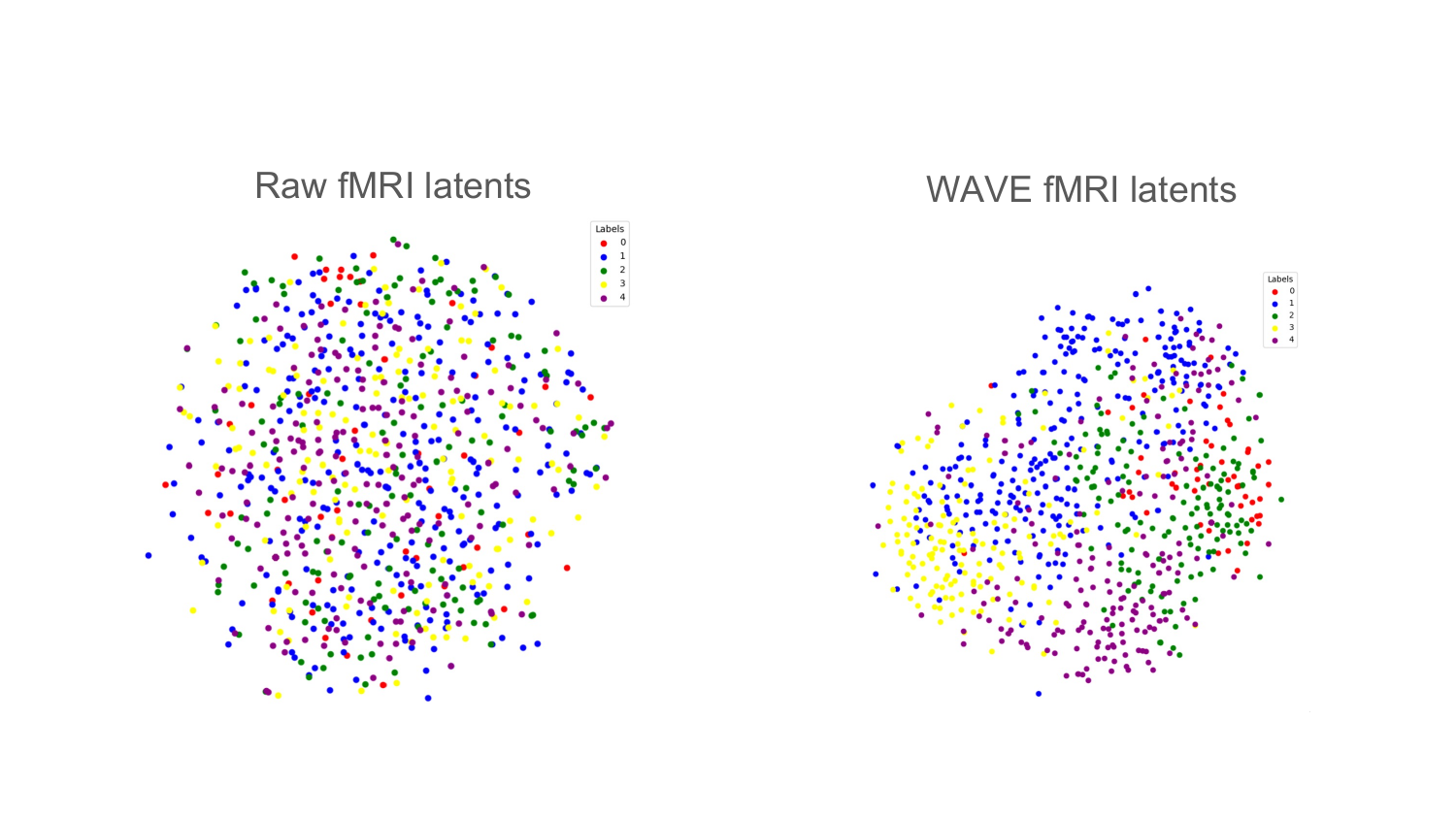}
    \vspace{-20pt}
    \caption{\textbf{T-SNE plot} depicting raw fMRI latents and WAVE encoded fMRI latents. The WAVE encoding method shows a clear defined separation into 5 clusters compared to the distribution of the raw latents.}
    \label{fig:fmri_latent}
\end{figure}

        \blue{Recent work has used linear methods like ridge regression for fMRI encoding~\cite{takagi2023high}. To evaluate our pretrained fMRI encoder, we conduct two ablation studies. First, we test fMRI-to-image retrieval using CLIP cosine similarity between projected fMRI and image embeddings (Table~\ref{tab:encoder_retrieval}). Second, we train encoders end-to-end on subject classification across all 4 subjects in BOLD5000 (Table~\ref{tab:subject_classification}). Both tasks demonstrate that (1) non-linear encoders substantially outperform linear baselines, and (2) pretraining provides significant additional gains, capturing both visual semantics and cross-subject variance.}
        \begin{table}[h]
\caption{\blue{\textbf{fMRI Encoder Ablation: Retrieval Accuracy.} We compare three fMRI encoder configurations on Brain-Vision and Brain-Language retrieval tasks. The pretrained WAVE encoder outperforms both the non-pretrained version and linear baseline, demonstrating that autoregressive pretraining learns robust fMRI representations for semantic alignment.}}
\vspace{5pt}
\begin{adjustbox}{width=\columnwidth}
\blue{
\begin{tabular}{l|cc}
\toprule
\textbf{Model} & \textbf{Brain-Vision Accuracy~(\%)} & \textbf{Brain-Language Accuracy~(\%)}                 
\\ \hline
pretrained~(WAVE) & \textbf{53.98} & \textbf{36.28} \\ \hline
w/o Pretrained & 50.22 & 28.84 \\ \hline
MLP backbone & 47.36 & 26.57 \\ \hline
Linear & 40.78 & 25.67 \\ 
\bottomrule
\end{tabular}
}
\end{adjustbox}
\label{tab:encoder_retrieval}
\end{table}
        \begin{table}[h]
\centering
\caption{\blue{\textbf{fMRI Encoder Ablation: Subject Classification Accuracy.} We evaluate each encoder's ability to capture cross-subject variance through subject classification on BOLD5000. The pretrained WAVE encoder achieves highest accuracy, showing that pretraining effectively learns cross-subject patterns in fMRI data.}}
\vspace{5pt}
\begin{adjustbox}{width=0.7\columnwidth}
\blue{
\begin{tabular}{l|c}
\toprule
\textbf{Model} & \textbf{Subject Classification~(\%)} \\ 
\hline
pretrained~(WAVE) & \textbf{84.78} \\ \hline
w/o Pretrained & 76.24 \\ \hline
MLP backbone & 74.85 \\ \hline
Linear & 68.32 \\ 
\bottomrule
\end{tabular}
}
\end{adjustbox}
\label{tab:subject_classification}
\end{table}
    \paragraph{fMRI Embedding.}
    In this study, we compare raw fMRI embeddings with our WAVE-encoded fMRI embeddings using a t-SNE visualization as depicted in Figure~\ref{fig:fmri_latent}. Typically, raw fMRI data visualized without encoding appears scrambled and indistinguishable across different clusters. In contrast, the WAVE-encoded fMRI embeddings clearly delineate distinctions between clusters. This distinction suggests that our WAVE model is effectively capturing the salient features of whole-brain fMRI signals.  
    \begin{figure}[t]
    \centering
    \includegraphics[width=0.5\linewidth]{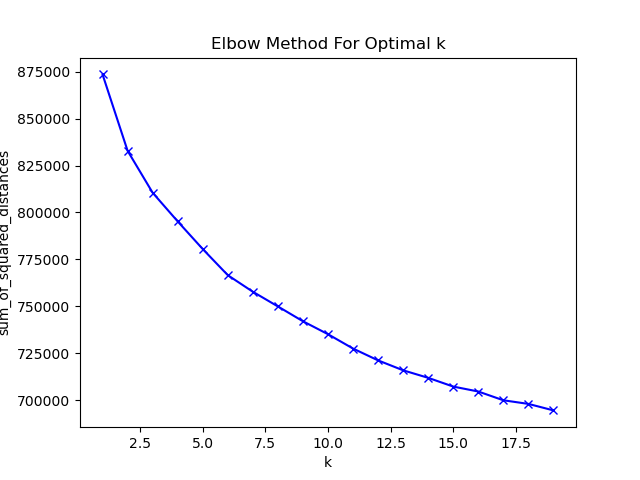}
    \vspace{-20pt}
    \caption{\textbf{Elbow Method for identifying the optimal number of clusters:} The reported figure is the sum of squared distance~(SSD) versus cluster number~$K$. While the SSD decreases as the number of clusters increases, there is no distinct ``elbow" point, which is difficult to determine an optimal cluster number from the curve.}
    \label{fig:elbow}
\end{figure}
    \subsection{Method}
    \label{app:method}
    \blue{
    \paragraph{fMRI Foundation Model.}
    To leverage large-scale neural representations, we utilize the fMRI foundation model developed by Thomas et al.~\cite{thomas2022self}, which employs a decoder-only Transformer architecture (GPT-2)~\cite{radford2019language}. To standardize inputs and capture whole-brain responses efficiently, fMRI data is parcellated into 1024 regions of interest using the DiFuMo1024 atlas~\cite{dadi2020fine}. This fine-grained soft parcellation preserves approximately 70\% of the explained variance in neural signals while reducing spatial dimensionality. Formally, an fMRI sequence is defined as $X \in \mathbb{R}^{T \times N}$, where $T$ represents time points and $N=1024$ represents the network parcels. During pre-training, the model treats each fMRI volume (TR) as a discrete token. Given different fMRI datasets have varies TRs, the fMRI foundation model treats TR as positional embeddings added to the fMRI tokens. These inputs are projected into an embedding space $E \in \mathbb{R}^{T \times D}$, where $D$ is the embedding dimension. The model is trained via a self-supervised autoregressive objective to predict the fMRI image at $T+1$ based on the history of previous fMRI images, effectively learning generalizable temporal brain dynamics. You can find details such as acquisition protocols in original paper. We initialize our encoder with these pre-trained weights\footnote{Code available at: \url{https://github.com/athms/learning-from-brains}} and fine-tune the model on the BOLD5000 dataset for the visual decoding task. 
    }
    \paragraph{Training Details.}
    For our training process, we consistently utilized a seed value of 42. We employed an Adam optimizer, setting the learning rate at $1 \times 10^{-4}$. The decay rates were configured differently for various components of the training: $0.03$ for the contrastive learning part and $0.02$ for the diffusion model training. The number of training steps was set to 50,000 for the contrastive learning phase and 200,000 for the diffusion model phase, for each individual model. In the case of universal model training, which involved data from four subjects, we quadrupled the training steps to accommodate the increased data volume. \blue{We also conducted hyperparameters search as described in Table~\ref{tab:training_hyperparams}}.
    \begin{table}[ht!]
\centering
\footnotesize
\caption{\blue{\textbf{WAVE Model Training Hyperparameters.} Configuration details for training phases across individual and universal model settings.}}
\renewcommand{\arraystretch}{1.2}
\begin{tabular}{lc}
\toprule
\textbf{Hyperparameter} & \textbf{Value} \\
\midrule
Random Seed & 42 \\
Optimizer & Adam \\
Learning Rate & Log-Uniform(5e-5, 1e-3) \\
\midrule
\textit{Weight Decay} & \\
\hspace{3mm} Contrastive Phase & Log-Uniform(1e-3, 5e-2) \\
\hspace{3mm} Diffusion Phase & Log-Uniform(1e-3, 5e-2) \\
\bottomrule
\end{tabular}
\label{tab:training_hyperparams}
\end{table}
    \paragraph{Compute.} 
        All experiments were conducted on single GPU setups. For the contrastive learning phase, we utilized Tesla V100 GPUs with 32GB of memory, where the WAVE models were trained for approximately 6 hours. In contrast, training the decoding diffusion model required significantly more GPU memory and computational power. We used an A-100 GPU with 40GB for this purpose, with the training duration for individual subject decoding approximately 25 hours. For the universal decoding model, which includes data from four subjects, the training time was extended to nearly 80 hours. The model inference also involves a diffusion model and we used one A-100 GPU. For fMRI preprocessing, we only utilized cpu compute nodes.
    
    \paragraph{Clustering.}
        The self-supervised clustering method that we used is k-clustering after the encoded image features by CLIP. It is not easy to determine how many clusters we should use in the analysis. BOLD5000 includes broad range of image types and classes, and we also conducted the elbow analysis for better cluster numbers. As we showed in Fig.~\ref{fig:elbow}, the sum of the squared distance is decreasing with larger cluster number. But there is no obvious elbow curve occurs even in 20 clusters. For better clustering analysis purpose, we used 5 clusters for our study. 


    \paragraph{Metrics.}
    For the evaluation and analysis of our model, we employed various metrics, including the mean cosine distance between scenario text and training images. This metric is computed as described in Equation~\ref{form:mean_consine_distance}. We provide the Python scipy code~\citep{virtanen2020scipy} used for calculating the mean cosine distance for each scenario texts, as follows:
    \begin{verbatim}
# Calculate cosine distance
from scipy.spatial.distance import cdist
cosine_distances = cdist(image_embeddings, 
                    _gt_text_embeddings, metric='cosine')
mean_cosine_distance = np.mean(cosine_distances)

\end{verbatim}

\end{document}